\documentclass{article}

\usepackage{microtype}
\usepackage{graphicx}
\usepackage{subfigure}
\usepackage{booktabs} % for professional tables

\usepackage{hyperref}

\usepackage[accepted]{icml2025}

\usepackage{amsmath}
\usepackage{amssymb}
\usepackage{mathtools}
\usepackage{amsthm}

\usepackage[capitalize,noabbrev]{cleveref}

\theoremstyle{plain}

\theoremstyle{definition}

\theoremstyle{remark}

\usepackage[textsize=tiny]{todonotes}

\icmltitlerunning{Statistical Scenario Modelling and Lookalike Distributions for Multi-Variate AI Risk}

\begin{document}

\twocolumn[
\icmltitle{Statistical Scenario Modelling and Lookalike Distributions for Multi-Variate AI Risk}



\icmlsetsymbol{equal}{*}

\begin{icmlauthorlist}
\icmlauthor{Elija Perrier}{yyy,comp}
% \icmlauthor{Firstname2 Lastname2}{equal,yyy,comp}
% \icmlauthor{Firstname3 Lastname3}{comp}
\end{icmlauthorlist}

\icmlaffiliation{yyy}{Centre for Quantum Software and Information, UTS, Sydney}
\icmlaffiliation{comp}{Stanford Center for Responsible Quantum Technology, Stanford University}

\icmlcorrespondingauthor{Elija Perrier}{elija.perrier@gmail.com}

\icmlkeywords{Risk, AI, Statistics, Scenarios}

\vskip 0.3in
]



\printAffiliationsAndNotice{\icmlEqualContribution} % otherwise use the standard text.

\begin{abstract}
Evaluating AI safety requires statistically rigorous methods and risk metrics for understanding how the use of AI affects aggregated risk. However, much AI safety literature focuses upon risks arising from AI models in isolation, lacking consideration of how modular use of AI affects risk distribution of workflow components or overall risk metrics. There is also a lack of statistical grounding enabling sensitisation of risk models in the presence of absence of AI to estimate causal contributions of AI. This is in part due to the dearth of AI impact data upon which to fit distributions. In this work, we address these gaps in two ways. First, we demonstrate how scenario modelling (grounded in established statistical techniques such as Markov chains, copulas and Monte Carlo simulation) can be used to model AI risk holistically. Second, we show how lookalike distributions from phenomena analogous to AI can be used to estimate AI impacts in the absence of directly observable data. We demonstrate the utility of our methods for benchmarking cumulative AI risk via risk analysis of a logistic scenario simulations.
\end{abstract}

\section{Introduction}
AI systems are increasingly integrated within multi-stage scenarios, supply chains, industrial workflows, or large-scale production pipelines. These deployments carry new risks regarding flow-on effects of AI use on other system components that are poorly understood or underestimated. Despite a burgeoning literature on AI risk and safety \cite{armstrong2012thinking,armstrong2018good,safelyinterruptibleagents,corrigibility}, most AI evaluation techniques focus only upon isolated model performance, such as classification accuracy, adversarial vulnerabilities \cite{ziegler2022adversarial,perez2022red}, \cite{ganguli2022red}, interpretability, misalignmnet or fairness in isolation \cite{lipton2018mythos}, rather than the effects of AI use across interconnected component systems. AI risk evaluations also typically lack a rigorous statistical grounding typical of conventional risk analysis \cite{miller2024adding}, including sampling strategies, distribution fitting, likelihood estimation and hypothesis testing of AI effects to be evaluated. Moreover, there also tends to be a lack benchmarking practices which sensitise models to enable risk metrics from non-AI workflows to be compared with AI-workflows. This is in part due to a dearth of source data on the measurable impacts of AI from which to perform statistical analysis. It is also due to difficulties disentangling what is uniquely AI risk from other non-AI related procedures.

While understanding model-specific risks is important, realistic deployment of AI involves its modular integration as one component of existing processes and risk frameworks. For AI risk and safety protocols to be of any practical use, it is therefore crucial to understand how the modular integrated use of AI, as a substitute for existing component, or additional component, changes the multi-variate risk profile of complex workflows and processes. In almost all practical use cases, the impact of an AI component failure is unlikely to be restricted to that component alone. As with any other technology, the effects of AI failure can be expected to propagate throughout workflows and processes. The failure of an AI component may have a significant follow-on effects on interconnected components, altering their profile, distribution and likelihood of failure. 
The use of AI across multiple components of a process may also lead to an accumulation of failures or produce unexpected side effects. Conversely, the use of certain AI components might reduce both total overall risk metrics and those of individual components within a workflow. It is therefore reasonable to assume that any practical analysis of the integrated AI must account for its use a component or module within a broader \textit{scenarios} - sequence of interdependent and interconnected events leading to a risk event.

\subsection{Contributions}
\label{sec:contrib}
To this end, our contributions are as follows:
\begin{enumerate}
    \item \emph{Scenario modelling of AI risk}. We demonstrate how to partition AI-related events into a multi-stage pipeline of sub-events $E_i$ whose properties (such as failure modes or loss) can be modelled as measurable random variables $X_i$ drawn from baseline distributions $F_i$. We then substitute $F_i^{(\mathrm{AI})}$ where AI is adopted, capturing the interplay of AI and non-AI components.
    \item \emph{Statistical empirical methods}. We illustrate how to deploy Markov chains and copulas to model dependencies among sequential failures and cumulative variation in risk metrics.
    \item \emph{Data scarcity solution via lookalike distributions}. We propose adapting distributions from comparable processes or partial automation scenarios to approximate $F_i^{(\mathrm{AI})}$ until sufficient AI-specific data are available.
    \item \emph{Benchmarking AI vs.\ non-AI process risks}. We show how scenario modelling enables side-by-side comparisons of baseline (no AI), partial AI, and full AI, revealing differences in probability of total overall loss or impact  $\Pr(X_{\text{total}} > t)$, mean performance, and catastrophic tail events.
    \item \emph{Example simulation}. We illustrate these methods on a three-stage pipeline (demand forecasting, warehouse picking, last-mile delivery) to demonstrate how scenario modelling allows pragmatic assessment of multi-variate AI risk.
\end{enumerate}

Overall, our framework aims to help operationalise AI risk in a manner aligned with established operational risk methodologies, bridging a gap in AI safety research and offering a replicable template for industry practitioners.

\section{Background}

\subsection{Related Work}

AI risk and safety is an emerging research discipline with particular importance to the deployment of AI systems. At one end of the spectrum, AI safety involves broad conceptual discussions of alignment and existential threats \cite{bostrom2014superintelligence, russell2019human}. This includes the risk of exceeding human oversight  \cite{moravec1988,vinge1993}, misalignment with human interests \cite{joy2000,yudkowsky2001} and catastrophic AI risk \cite{hendrycks2023overview}. At the other, more technically focused end, AI safety examines specific model behaviour and artefacts that can give rise to adverse impacts. This includes substantial work on adversarial robustness \cite{viano2021robust,croce2020robustbench}, interpretability \cite{lipton2018mythos}, and distribution shift \cite{quinonero2009dataset,koh2021wilds,wiles2021fine}. Yet despite some discussion of how AI and non-AI components may interact (see \cite{hendrycks2025introduction}), these approaches often focus upon single models or datasets, rather than on the wider system-level or multi-stage nature of deployment. 

By contrast, existing risk disciplines, such as operational risk modelling, particularly in finance and insurance, applies rigorous statistical tools—such as heavy-tailed distributions, dependence structures, and scenario-based approaches—to analyse correlated losses across interconnected processes \cite{mcneil2015quantitative,embrechts2002correlation}. Likewise, reliability engineering has developed a rich set of formal methods (e.g.\ Markov chains, renewal processes) for predicting time-to-failure in multi-component systems \cite{rossstochastic,coles2001introduction}. These mature techniques have only recently begun to surface in AI risk contexts. Yet the approaches are piecemeal at best, with AI safety yet to be premised upon anything like a science of AI risk. Several works discuss or imply multi-event risk frameworks for AI. Some AI safety perspectives note that failures do not occur in isolation, and instead emphasise practical or contextual considerations \cite{amodei2016concrete} and operational embedding \cite{varshney2019risk}. Other literature employs discrete-event simulations or extended forecasting models but do not specifically foreground AI risk. In recent literature, we have seen emerging trends for incorporating statistical methods traditionally seen in finance or reliability theory directly into AI risk analysis, evaluations \cite{miller2024adding}, exploring Markov-based reliability, copula-based tail dependence \cite{joe2014dependence}, scenario-based hazard analyses, and even system-level threat models adapted to emergent vulnerabilities.

Despite these advances, few holistic studies systematically integrate copulas, Markov processes, mixture models, or heavy-tail corrections into \emph{multi-stage} AI risk scenarios. In particular, the idea of combining lookalike distributions for data-scarce AI components with domain-informed parameter shifts has received limited attention beyond preliminary concepts set out in standard texts \cite{mcneil2015quantitative}, and a handful of scenario-based enterprise risk frameworks. As a result, many real-world AI deployments lack robust quantitative estimates for tail outcomes, correlated component failures, or potential cascading effects. Our work contributes to closing this gap by merging established statistical risk techniques with explicit recognition of AI’s emerging failure profiles in multi-stage workflows.

\subsection{Statistical Scenario Modelling}
 Scenario modelling is a technique widely used in operational risk analysis and reliability theory. It offers a systematic method for evaluating how the inclusion of AI components in a multi-stage workflow alter overall risk, including robust assessments of correlated losses and catastrophic events \cite{mcneil2015quantitative,embrechts2002correlation,rossstochastic}. The core idea behind scenario modelling for AI is to decompose complex workflows that utilise AI into sub-events. Data together with theoretical methods are used to quantifies how each sub-event behaves statistically by estimating each sub-event or component distributions. A dependency relation is chosen in order to interconnect the distributions in ways that enable estimation of risk measures, such as value at risk or time to failure (including catastrophic AI risk \cite{russell2019human,testimony_russell,tegmark2018life}). Dependency relations between components are typically modelled using Markov chains and copulas.  Monte Carlo simulations are then used to extract statistics upon which risk metrics are based. 

One of the benefits of scenario modelling for AI risk analysis is that it provides a means of  \emph{sensitising} workflows by introducing distributions that capture the performance of AI components at each stage. In the logistics example we study below, we assign one distribution to human-driven warehouse picking and another to AI-driven robotics picking, and then compare risk metrics of other components and overall to assess the variation in risk arising from the use of AI. Sensitising scenarios is crucial for organisational decisions as to whether and how to integrate AI into their workflows. Scenario modelling is a natural tool for doing so.

\subsection{Event Sequencing}
\label{sec:scenario_modelling}
Scenario modelling relies upon decomposing an event, such as the chance of a system failure due to AI, into a sequential pipeline of component sub-events. We consider a pipeline as a directed graph of events 
$E_{1} \to E_{2} \to \cdots \to E_{n}$.
Each event $E_i$ is associated with a random variable $X_i$ drawn from a distribution $F_i$ that encodes a relevant outcome of a component in the process, such as cost, delay, or a risk measure at that stage in the workflow. In a baseline (non-AI) scenario, $X_i \sim F_i$, where $F_i$ is estimated from historical data or domain knowledge. Introducing AI at event $E_i$ modifies that distribution to $F_i^{\mathrm{(AI)}}$. This can lower average errors or processing times while also shifting tail behaviour, for instance by adding a small but increased chance of a catastrophic failure.
We let the aggregate system outcome be either the sum:
\begin{align}
  X_{\mathrm{total}} &= \sum_{i=1}^n X_i
\end{align}
or a more general function:
\begin{align}
  X_{\mathrm{total}} &= g\bigl(X_1, X_2, \ldots, X_n\bigr),
\end{align}
depending on how sub-event outcomes interact. In our examples below, $X_i$ might be some measure of cost, adverse impact or loss. Events $E_i$ and their impacts $X_i$ are often correlated in complex ways. This means that failure in one stage may directly impact subsequent stages or that exogenous factors may affect multiple stages simultaneously. Thus assessing the aggregate impact of events - and the integration of AI within processes - is rarely straightforward. As noted above, doing so ideally involves understanding the interconnected dependencies between events. Markov chains for example can be used to capture these dependencies when the system passes through discrete states like operational, minor-failure, and repair. Copulas are useful when continuous variables have correlated or heavy-tailed joint behaviour. We discuss both dependency relations below.

\subsection{Data scarcity}
\label{subsec:data_scarcity}
Being able to model event dependencies relies upon having data or information from which to estimate relevant distributions (or probabilities) of those events occurring. Scenario modelling works most effectively where there is an abundance of observational data that allows for rigorous fitting of distributions (such as data obtained via measurement of production processes or workflows). A major challenge in AI-integrated workflows is the lack of extensive historical data to reliably estimate AI-specific failure distributions. This contrasts with traditional automation or mechanical systems, for which well-catalogued failure times or defects allow direct parameter fitting. To address this problem, we present two approaches to constructing \emph{lookalike distributions} that serve as proxies for AI risk until more direct AI data becomes available: (i) AI-specific tail adjustments and (ii) lookalike distributions with goodness-of-fit updates. We demonstrate each using our logistics workflow example. 

\subsubsection{AI-specific tail adjustments.}
Suppose that in a three-stage logistics pipeline, the warehouse picking task $E_2$ is replaced by an AI-driven robotic system. If we have a known mechanical-automation distribution from the literature, say a Gamma distribution \cite{snyder1984inventory}:
\begin{align}
  \widetilde{T} \;\sim\; \Gamma\bigl(\alpha, \beta\bigr),
\end{align}
where \(\alpha\) (shape) and \(\beta\) (scale) were estimated from non-AI robotic picking data, we may use this as a starting point. To reflect new AI-specific risk, we can adjust the tail to allow for a small probability \(\varepsilon\) of extreme failures:
\begin{align}
  T_{\mathrm{AI}}
  \;=\;
  \begin{cases}
    \widetilde{T}, \quad & \text{with probability }(1-\varepsilon),\\
    Z, \quad & \text{with probability }\varepsilon,
  \end{cases}
\end{align}
where \(Z\) is drawn from a heavy-tailed distribution such as a Generalised Pareto distribution with shape parameter \(\xi > 0\). This yields a mixture model:
\begin{align}
  F_{T_{\mathrm{AI}}}(t)
  \;=\;
  (1-\varepsilon)\,F_{\widetilde{T}}(t) \;+\; \varepsilon\,F_{Z}(t).
\end{align}
Modelling the body and tail separately via a mixture model is widely discussed in the extreme value theory and operational risk literature \cite{coles2001introduction, embrechts1997modelling, castellanos2007default, sy2021extreme, wang2023generalized}. In practice, \(\varepsilon\) and \(\xi\) can be chosen using domain-expert judgment about AI’s susceptibility to rare edge-case failures, which may increase tail heaviness beyond what mechanical systems exhibit. Over time, if actual AI picking data indicates fewer or more extreme incidents, \(\varepsilon\) or \(\xi\) can be recalibrated.

\subsubsection{Lookalike plus goodness-of-fit updates}
An alternative is to assume that AI initially follows the same distribution as the non-AI component, then iteratively update parameters as AI usage data becomes available. For example, suppose the pre-AI system’s picking times follow
\begin{align}
   \widehat{T} \;\sim\; \mathrm{Weibull}\bigl(k, \lambda\bigr)
\end{align}
a distribution commonly used in reliability analysis \cite{meeker1998statistical}. In the absence of direct AI data, we can initially set:
\begin{align}
   T_{\mathrm{AI}} \;\sim\; \mathrm{Weibull}\bigl(k^{(0)}, \lambda^{(0)}\bigr),
\end{align}
using the same \(\bigl(k, \lambda\bigr)\) or minor adjustments \(\bigl(k^{(0)}, \lambda^{(0)}\bigr)\) \cite{dixon1997iterative}. As AI picking data \(\{x_1, x_2, \dots, x_n\}\) is obtained, we re-estimate parameters by maximum likelihood:
\begin{align}
   \bigl(\hat{k}, \hat{\lambda}\bigr)_{\mathrm{new}}
   \;=\;
   \mathrm{arg\,max}_{k,\lambda}
   \prod_{i=1}^n 
     f_{\mathrm{Weibull}}(x_i \mid k,\lambda),
\end{align}
or use a Bayesian update if we have prior distributions on \(k,\lambda\) \cite{beck1977maximum}. After refitting, we check whether the Weibull form remains appropriate by applying a goodness-of-fit test (for instance, a Kolmogorov–Smirnov or Anderson–Darling test) \cite{rausand2004risk}. If repeated experimental data suggests a poor fit, we can consider a heavier-tailed family or mixture model to capture AI-specific anomalies. This iterative approach treats the non-AI component as a \emph{lookalike} baseline and adapts it using real operational data from the AI-driven system. In both methods, these AI-related distributions can be inserted into the scenario modelling framework described in Section~\ref{sec:scenario_modelling}, replacing \(F_i\) by \(F_i^{(\mathrm{AI})}\). As more AI data accumulate, parameters can be continuously revised. Such iterative refinement is crucial for capturing novel failure modes and for highlighting whether AI’s true risk profile deviates from initial assumptions about mechanical or purely human-driven processes. A third approach, which we do not discuss here is to use machine learning methods to either learn the distribution or, where knowledge of the distribution may be impractical or intractable to estimate, to learn key features of the distribution relevant to control and mitigation.

\subsection{Dependency Structures}
\label{subsec:theory}
Multi-event scenario modelling requires a means of interconnecting each event distribution $F_i$ into an overall scenario evolution. Two commonly used methods for doing so are Markov Chains and copula methods together with Monte Carlo simulations \cite{coles2001introduction}. Markov chains suit event pipelines where each sub-event is modelled in terms of states whose configurations, such as normal operation, minor failure, or critical failure, can be modelled via probability distributions. These distributions are used to estimate the likelihood of the component entering into a particular failure mode (or sub-event).  Alternatively, a copula-based approach can capture dependencies among continuous sub-event distributions. Scenario modelling provides a way to assess how AI failure modes propagate through systems. This is can be especially useful partial automation or full automation changes the correlation structure among components \cite{joe2014dependence,coles2001introduction}.
\subsubsection{Markov Chains}
Markov chains are widely used for modelling state-based transitions, commonly encountered in reliability engineering (operational/failure/repair states) \cite{rossstochastic}. If we assume a discrete-time Markov chain, then at each time step $t$, the system occupies a state $X_t$ from a finite set $\{0,1,2,\dots\}$. The transitions obey:
\[
  p_{ij} = \Pr(X_{t+1} = j \mid X_t = i)
\]
and $\sum_j p_{ij} = 1$. In the context of AI risk, a state might represent different operational modes (e.g., normal operation, partial failure, major failure, under-repair). Substituting AI may change specific transition probabilities. For instance, the probability of going from normal operation to partial failure could decrease (better performance on average), but the probability of jumping directly to major failure might be introduced or increased if the AI fails in an unanticipated way. A continuous-time Markov chain variant can also be used by defining exponential rates $\lambda_{ij}$. The choice depends on whether the sub-event times are best approximated in discrete steps or continuous durations.

\subsubsection{Copulas}
When the primary random variables of interest (e.g., time to completion, cost, or quantity of defects) are continuous, a copula-based approach helps capture correlations and tail dependencies \cite{joe2014dependence}. Copulas are also of interest in statistics as they provide a means of assessing scale-free dependence and for use in simulation \cite{nelson2006introduction}. Sklar’s theorem states that for any joint distribution $F_{X_1,\dots,X_n}$, there exist marginal distributions $F_{X_i}$ and a copula $C$ such that:
\[
   F_{X_1,\dots,X_n}(x_1,\dots,x_n) 
   = C\bigl(F_{X_1}(x_1),\dots,F_{X_n}(x_n)\bigr).
\]
By introducing an AI-specific distribution $F_i^{\mathrm{(AI)}}$ or altering copula parameters to reflect new correlation patterns, the resulting model shows how a modest improvement or degradation in one stage’s reliability can ripple across the entire chain. We can thus separately choose each marginal $F_{X_i}$ to fit sub-event data and then select a copula (Gaussian, t-copula, or Archimedean families like Clayton or Gumbel) to encode dependence. If AI modifies sub-event $E_i$, we replace $F_{X_i}$ with a new distribution or update the copula parameter if the correlation structure is believed to shift.

\subsubsection{Monte-Carlo Simulations}
Scenario models typically rely upon simulations for which Monte Carlo methods are the most common method. Each simulation run draws one sample from each sub-event distribution, subject to the chosen dependence model and then computes a system-wide outcome (for instance, total delay or cost). Repeating this many times yields an empirical distribution of outcomes and measurement statistics, from which decision-relevant metrics can be derived. these include the probability of exceeding certain thresholds, the Value at Risk (VaR) for a chosen confidence level, or the Expected Shortfall of catastrophic events \cite{mcneil2015quantitative,robert2004monte}. Such metrics can indicate whether AI adoption in a particular scenario merely shifts risk profiles or genuinely mitigates overall hazards. They can also illuminate potential trade-offs between typical gains and the possibility of rare but extreme failures. We discuss these below.

\subsection{Risk Measures}
\label{subsec:quant_risk_analysis}
As noted above, the purpose of scenario modelling for AI is to enable statistically meaningful quantification of how risk metrics vary under a range of conditions due to the integration of AI. Risk measures in operational risk settings aim to capture the likelihood or severity of adverse outcomes. Two widely used measures are \textit{Value at Risk (VAR)} and \textit{Expected Shortfall (ES)} (also known as conditional VAR).

\subsubsection{Value at Risk (VaR)}
Value at Risk is given by:
\begin{align}
   \mathrm{VaR}_{\alpha}(L) = \inf \{ \ell : \Pr(L \le \ell) \ge \alpha\}
\end{align}
where $L$ is the loss random variable and $\alpha$ is a specified confidence level (e.g. 0.95, 0.99). $\mathrm{VaR}_{\alpha}(L)$ is a threshold that the loss does not exceed with probability $\alpha$. It is a popular measure in finance \cite{jorion2006value, artzner1999coherent} but is sometimes criticised for ignoring the shape of the distribution beyond the chosen range.

\subsubsection{Expected Shortfall (ES)}
Expected shortfall is given by:
\begin{align}
   \mathrm{ES}_{\alpha}(L) = \frac{1}{1-\alpha}\int_{\alpha}^{1}\mathrm{VaR}_{u}(L)\,du
\end{align}
which captures the expected (or mean) loss in the worst $(1-\alpha)$ fraction of cases. ES is considered more sensitive to tail behaviour than VaR and can give a clearer picture of catastrophic outcomes \cite{acerbi2002expected, rockafellar2000optimization}. Other domain-specific thresholds or cost-based metrics (e.g., probability that downtime exceeds 24 hours) can also be used. 

\subsection{Sensitising AI Scenarios}
The extent of AI integration in any workflow or process is a matter of degree. In almost all cases, AI is simply a sub-component of a particular process that involves non-AI components (albeit advances in frontier and LLMs are seeing advances in capability that allow AI to construct and execute across entire pipelines). To reliably analyse AI risk of any variety, it is useful where possible to benchmark the use of AI against an equivalent or near-equivalent process using sensitivity analysis \cite{saltelli2008global,pianosi2015sensitivity,iooss2015review}. Doing so allows the application of statistical methods that can test for whether the integration of AI gives rise to any statistically significant differences in risk measures, such as those identified above. Benchmark comparisons are also useful as a means of identifying or detecting AI-related events themselves. In the simulations below illustrating our methods, we introduce three cases against which to test a scenario:
\begin{enumerate}
    \item \emph{Non-AI (Baseline)}. The system uses conventional processes, with established or historical data for each sub-event. We can typically model these distributions more confidently, given prior logs or well-understood reliability metrics.  All events remain as historically modelled, e.g., 
\begin{align}
   \{F_1, F_2, \dots, F_n\}.
\end{align}
    \item \emph{Partial AI}. AI replaces or augments only some sub-events. For example, the warehouse picking process might be automated, but forecasting and delivery remain unchanged. Suppose AI is introduced only for sub-event $E_k$, so that event has distribution 
\[
   F_k^{\mathrm{(AI)}},
\]
while the others remain $F_i$. If multiple events adopt AI, we selectively swap in their AI distributions.
    \item \emph{Full AI}. AI replaces all feasible sub-events. This scenario tests a maximum AI adoption strategy. All sub-events adopt the corresponding AI distributions:
\[
   \{F_1^{\mathrm{(AI)}}, F_2^{\mathrm{(AI)}}, \dots, F_n^{\mathrm{(AI)}}\}.
\]
\end{enumerate}
By comparing the distribution of outcomes (e.g., total cost, time, or the presence of catastrophic failure) across these three configurations, we can more easily isolate the net effect of AI adoption within margins of error.

\section{Methods}
We demonstrate the use of scenario modelling and lookalike distribution methods using a toy model of a logistics workflow process. We adopt the following process based on the scenario modelling and lookalike distribution principles articulated above for AI risk is set out below.
\begin{enumerate}
    \item \emph{Decompose the pipeline.} Identify sub-events $E_1, E_2, \ldots, E_n$, each with a random variable $X_i \sim F_i$ below. 
    
    \item \textit{Baseline Distributions}. Assign a baseline distribution $F_i$ (such as Gamma, Weibull, or Bernoulli) to characterise the event’s cost, time, or failure frequency. This is done ideally using empirical methodology and data to estimate suitable $F_i$, but it can also draw upon results from the literature.

    \item \emph{Estimate Lookalike Distributions}. When an AI module replaces sub-event $E_i$, change $F_i$ to $F_i^{\mathrm{(AI)}}$. This might reduce the mean but introduce heavier tails. In Markov settings, the transition probabilities (or rates) are also modified as a result. In a copula model, the relevant marginal or dependence parameters may shift to reflect new correlations.

    \item \emph{Set scenario model dependencies}. Use a Markov chain when sub-events or system states are discrete. If sub-events produce continuous outcomes, use a copula $C$ to link the marginal distributions $F_1,\ldots,F_n$. For example, in a Gaussian copula, one estimates the correlation matrix among $\Phi^{-1}(F_i(x))$ terms, where $\Phi$ is the standard normal CDF.

      \item \emph{Measure risk metrics}. After specifying distributions, we run Monte Carlo simulations to compute or simulate $  X_{\mathrm{total}} 
    $
    and evaluate metrics such as 
    $
      \Pr(X_{\mathrm{total}} > T)
    $,
    or the following classical measures:
    \begin{align}
      \mathrm{VaR}_{\alpha}(X_{\mathrm{total}})
      &= \inf\!\bigl\{\, x : \Pr\bigl(X_{\mathrm{total}} \le x\bigr) \ge \alpha \bigr\},\\
      \mathrm{ES}_{\alpha}(X_{\mathrm{total}})
      &= \frac{1}{1 - \alpha}\,\int_{\alpha}^{1}\mathrm{VaR}_{u}(X_{\mathrm{total}})\,du.
    \end{align}
    These highlight both moderate and tail-based risks.

    \item \emph{Compare scenarios}. The procedure above is implemented for three cases: no AI, partial AI, and full AI (where more events are substituted with AI). By contrasting distributions and risk metrics across these scenarios, one can assess whether AI integration decreases mean system losses or inadvertently increases low-probability, high-severity events.
\end{enumerate}

This iterative scenario modelling not only provides an initial forecast of AI-induced risk shifts but also forms a basis for ongoing updates. As more operational data on AI performance are collected, parameters can be re-estimated, enabling continuous monitoring of how AI affects system-wide risk.

\subsection{Example: Hypothetical Multi-Stage Logistics Pipeline}
\label{subsec:logistics_example}
We illustrate how scenario modelling can reveal the net effect of AI in a hypothetical supply chain pipeline with three sub-events: 
\emph{($E_1$) Demand Forecasting}, 
\emph{($E_2$) Warehouse Picking}, 
\emph{($E_3$) Last-Mile Delivery} where each process has a non-AI and AI instance.\\

\emph{$E_1$: Demand Forecasting.} 
\begin{itemize}
\item Non-AI baseline: A simplistic statistical or manual forecast approach. Historical data is used to model distribution of forecast errors, possibly Gaussian-like with moderate variance.
\item AI-based scenario: A deep learning forecast system with improved average accuracy but an uncertain tail that might produce large outliers. The distribution might show narrower bulk but heavier extremes.
\end{itemize}

\emph{$E_2$: Warehouse Picking.} 
\begin{itemize}
\item Non-AI baseline: Manual or semi-automated picking. The time distribution is known, with some moderate correlation to forecast accuracy.
\item AI-based scenario: A robotics-based system with machine vision. This might reduce mean picking time but introduce a small chance of catastrophic malfunction if the AI fails to identify items correctly.
\end{itemize}

\emph{$E_3$: Last-Mile Delivery.}
\begin{itemize}
\item Non-AI baseline: Traditional trucks or human couriers, known average times with a certain tail for traffic disruptions.
\item AI-based scenario: Drone delivery or autonomous vehicles. Potentially faster on average but with novel tail risks (e.g., navigation glitch).
\end{itemize}

We can examine partial AI (replacing only $E_1, E_2$ or $E_3$) or full AI (all three replaced). The scenario modelling approach draws from either discrete or continuous distributions for sub-events, plus a mechanism (Markov chain or copula) to handle correlations.

\subsection{Setting and Parameters}

\subsubsection{Baseline (Non-AI)}
Assume the pipeline processes shipments daily, with each day representing a new run of the three-step chain. We choose baseline distributions drawn from the literature as an example:
\begin{itemize}
\item $E_1$: Forecast error distribution is approximate normal with mean 0, standard deviation 15 (representing percent error). This leads to occasional overstock or understock.
\item $E_2$: Picking times follow a Gamma distribution with shape $k=5$ and scale $\theta=2$, giving a mean of $10$ (time units).
\item $E_3$: Delivery times are lognormal with parameters $(\mu=1.0,\sigma=0.5)$, yielding an average around $3.0$ but a long right tail.
\end{itemize}
For forecasting error, normal distributions are often used \cite{hyndman2018forecasting}. For travel time and time to destination, lognormal distributions are sometimes chosen due to their relative simplicity \cite{clark2005modelling}. And for picking times the Gamma distribution is also seen in the literature. With our baseline distributions now chosen, we incorporate moderate correlation via a Gaussian copula between $E_1$’s forecast error and $E_2$’s time, reflecting that large understock or overstock can slow picking. $E_3$ is assumed partially correlated with $E_2$ but with lesser strength.  

\subsubsection{Partial AI} Suppose we replace $E_2$ alone with an AI-driven robotics system. For example we might  hypothesise that  $E_2$’s distribution changes to $\Gamma(k=8,\theta=1)$, indicating an improved mean picking time of $8$. However, if the AI system fails, it fails drastically: we add a 2 percent chance that $E_2$ time is extremely large, modeled by a heavy tail or mixture distribution. The correlation to $E_1$ might also shift if forecast errors no longer hamper manual picking in the same way. For instance, the AI picking system might be less impacted by forecasting misalignments, partially decoupling $E_1$ from $E_2$.

\subsubsection{Full AI} For the case where AI is substituted for each event, we might consider $E_1$ being replaced by a deep learning model that significantly reduces average forecast error to standard deviation $10$, but introduces a small probability of a large error. $E_3$ may be substituted 
using autonomous drones or vehicles, cutting average delivery time but with new tail risks (a chance of a major route glitch or regulatory hold-up). We can adjust the copula parameters to reflect possible new correlations among $E_1, E_2, E_3$ if, for example, AI-based forecasting errors degrade the performance of AI-based picking or routing in correlated ways.

\subsection{Simulation Procedure}
\label{subsec:simulation}
To illustrate our method, we describe two simulations to capture dependencies among sub-events and generate multi-stage risk samples. Code for our simulations is available via an online repository \cite{reporisk}. Both start by specifying baseline (non-AI) and AI variants of $E_1$, $E_2$, and $E_3$, then combine them under a common framework (copula or Markov chain).

\subsubsection{Distributions}
We first assign marginal distributions. Let $E_1$ (forecasting) in the non-AI scenario have 
\begin{align}
\text{$E_1$: Non-AI} \qquad \; X_1\sim\; \mathcal{N}(0,\,15^2),
\end{align}
where forecasting errors are normally distributed with mean 0 and standard deviation 15. An AI-based forecaster uses a mixture model,
\begin{align}
X_1^{(\mathrm{AI})} \;=\;
\begin{cases}
\mathcal{N}(0,10^2), & \text{with probability }1-\varepsilon_1,\\
\text{Lognormal}(1,0.8), & \text{with probability }\varepsilon_1
\end{cases}
\end{align}
where $\varepsilon_1 \approx 0.02$ is a small probability mass accounting for extreme outliers. Here we have assumed that an AI forecaster has reduced the error to $\mathcal{N}(0,10^2)$ for most cases. However, the corollary we assume here is that for a small number of (for example) out of distribution forecast, forecast errors are considerable. The heavy-tailed lognormal is added to reflect occasional but extreme forecasting errors. We choose parameters $\mu = 1, \sigma = 0.8$ (with mean $\exp(1 + 0.8^2/2) \approx 3.74$) which indicates that when error does occur, it is large compared to the revised forecasting errors. 

For $E_2$ (picking), the non-AI version follows:  
\begin{align}
X_2 \sim \Gamma(5,2),
\end{align}
and the AI version:
\begin{align}
X_2^{(\mathrm{AI})} \;=\;
\begin{cases}
\Gamma(8,1), & \text{with probability }1-\varepsilon_2,\\
\mathrm{Pareto}(\alpha_2), & \text{with probability }\varepsilon_2.
\end{cases}
\end{align}
We assume that an AI-driven picking system ought to reduce average picking times, but that occasional system faults or extreme operational disruptions may cause drastic delays due complex nature of warehouse organisation and that picking usually lacks redundancy (the correct item must be picked). A Pareto distribution, with its characteristic power-law decay, is chosen to capture such tail risks.

Finally, $E_3$ (delivery) uses a lognormal baseline:
\begin{align}
X_3 \sim \mathrm{Lognormal}(1,\,0.5)
\end{align}
and for AI-based delivery,
\begin{align}
X_3^{(\mathrm{AI})} \;=\;
\begin{cases}
\mathrm{Lognormal}(0.8,0.4), & \text{with probability }1-\varepsilon_3,\\
\mathrm{Weibull}(k_3,\lambda_3), & \text{with probability }\varepsilon_3.
\end{cases}
\end{align}
In principle, the parameters $\varepsilon_i$, $\alpha_2$, $k_3$, or $\lambda_3$ can be updated as new data on AI performance emerge according to a chosen optimisation method (itself amenable to machine learning). Appendix \ref{Appendix: Lookalike Distribution Simulation} sets out a sketch of a potential simple method for doing so using LLMs.

\subsubsection{Copula-Based Sampling}
\label{subsubsec:copula_sampling}
To encode dependencies, we choose a Gaussian copula $C_{\Sigma}$ with correlation matrix $\Sigma$. If AI modifies correlations (e.g., decoupling picking times from forecasting errors), we adjust $\Sigma$ accordingly. For each Monte Carlo sample, we draw $(u_1, u_2, u_3)\;\sim\;C_{\Sigma}$ and invert $u_i$ through the relevant marginal (non-AI or AI). Summing or otherwise aggregating the resulting $(x_1, x_2, x_3)$ gives $X_{\text{total}}$. Repeating many samples yields an empirical distribution from which we compute risk metrics such as 
\(
\Pr(X_{\text{total}}>t)
\), 
VaR, or ES.

\subsubsection{Markov Chain Sampling}
\label{subsubsec:markov_sampling}

In a Markov chain approach, each sub-event transitions among discrete states (e.g.\ $\{\text{Operational, MinorFail, MajorFail}\}$). For the non-AI case, define a transition matrix $P_{\mathrm{nonAI}}$ with entries
\begin{align}
p_{ij}^{(\mathrm{nonAI})}
\;=\;
\Pr\bigl(X_{t+1}=j \,\mid\, X_t=i\bigr),
\end{align}
estimated from historical baselines. Introducing AI modifies certain transitions by a small shift $\Delta_{ij}(\mathrm{AI})$, resulting in 
\begin{align}
p_{ij}^{(\mathrm{AI})}
\;=\;
p_{ij}^{(\mathrm{nonAI})} \;\pm\;\Delta_{ij}(\mathrm{AI}),
\end{align}
reflecting lower probabilities for minor failures but a new (small) route to catastrophic failure. If $E_1$ is replaced by AI forecasting, then $E_2$’s transition probabilities might also shift if picking states depend on the forecasting outcome.

In simulation, one initialises the chain in an operational state and evolves it through the stages $E_1$, $E_2$, $E_3$. At each stage, we sample a state transition using either $p_{ij}^{(\mathrm{nonAI})}$ or $p_{ij}^{(\mathrm{AI})}$, depending on which sub-events adopt AI. We then accumulate cost/time/loss variables, for example by assigning a random cost distribution conditioned on the current state. After running the chain to completion, we obtain a total outcome $X_{\text{total}}$. Repeating over many runs builds an empirical distribution from which we derive probabilities of large losses, VaR, or other performance measures. As real AI data arrives, parameters $p_{ij}^{(\mathrm{AI})}$ or the cost distributions associated with each AI-related state can be updated. 

\section{Results}
\subsection{Hypothetical Outcomes}
Results for the hypothetical scenario using the copula method and Markov chain method are set out in Tables \ref{tab:copula_scenarios} and \ref{tab:markov_chain_scenarios} respectively. 

\begin{table}[h!]
\centering
\begin{tabular}{lrrrr}
\hline
\textbf{Scenario} & \textbf{Mean} & \textbf{P1} & \textbf{VaR(0.95)} & \textbf{ES(0.95)} \\
\hline
Non-AI    & 12.97 & 0.17 & 42.84 & 50.45 \\
Partial-AI & 10.86 & 0.11 & 37.78 & 44.82\\
Full-AI    & 11.83 & 0.05 & 30.58 & 63.09 \\
\hline
\end{tabular}
\caption{Copula-Based Scenario Results. P1 is P($X_{\text{total}}>30$).}
\label{tab:copula_scenarios}
\end{table}

\begin{table}[h!]
\centering
\begin{tabular}{lrrrr}
\hline
\textbf{Scenario} & \textbf{Mean} & \textbf{P1} & \textbf{VaR(0.95)} & \textbf{ES(0.95)} \\
\hline
Non-AI    & 39.57 & 0.04 &  91.44 & 129.83 \\
Partial-AI & 40.36 & 0.06 & 127.64 & 165.64 \\
Full-AI    & 43.94 & 0.11 & 166.60 & 210.77 \\
\hline
\end{tabular}
\caption{Markov Chain Scenario Results. Here $P1$ is P($X_{T} > 100$).}
\label{tab:markov_chain_scenarios}
\end{table}

\begin{figure*}[ht]
    \centering
    % First sub-figure
    \subfigure[Copula Model Results]{
        \includegraphics[width=0.45\textwidth]{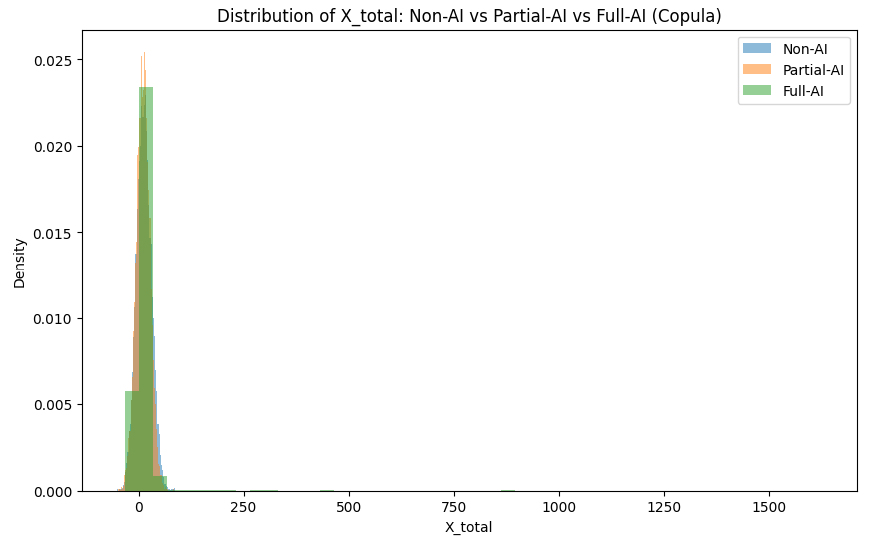}
        \label{fig:copula}
    }
    \hfill
    % Second sub-figure
    \subfigure[Markov Chain Results]{
        \includegraphics[width=0.45\textwidth]{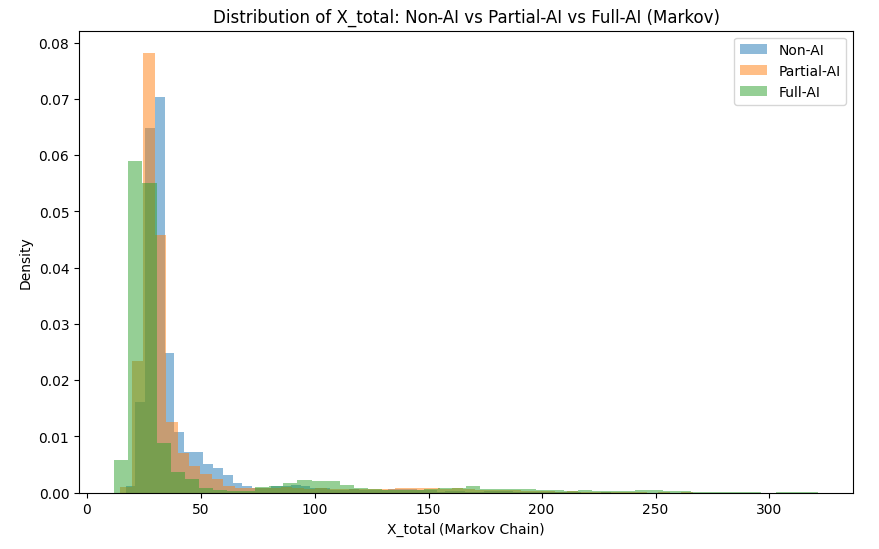}
        \label{fig:markov}
    }
    \caption{(a) Distribution of $X_{\mathrm{total}}$ under Non-AI, Partial-AI, and Full-AI scenarios using a Gaussian copula model. (b) Distribution of $X_{\mathrm{total}}$ under Non-AI, Partial-AI, and Full-AI scenarios using the Markov chain approach. }
    \label{fig:side_by_side}
\end{figure*}

\section{Discussion}

Figure~\ref{fig:side_by_side}(a) and Table~\ref{tab:copula_scenarios} present a copula-based analysis of how $X_{\mathrm{total}}$ (total cost or time) is distributed under Non-AI, Partial-AI, and Full-AI. In Figure~\ref{fig:side_by_side}(a), the Partial-AI case reduces both the mean and moderate tail probabilities compared to the Non-AI case, yet the Full-AI scenario, while potentially improving median performance, exhibits a heavier tail. Table~\ref{tab:copula_scenarios} confirms this by displaying a lower probability $P(X_{\mathrm{total}}>30)$ for Full-AI but a significantly higher Expected Shortfall (ES), underscoring the trade-off between average improvement and catastrophic failure risk.

Figure~\ref{fig:side_by_side}(b) and Table~\ref{tab:markov_chain_scenarios} show analogous results for a Markov chain approach, which models discrete operational states and transitions. Although the Partial-AI scenario shows a similar mean as Non-AI, it leads to a larger Value-at-Risk (VaR) and ES, indicating heavier tails once failures propagate. The Full-AI scenario pushes these tail metrics even higher, demonstrating how correlated AI failures across multiple stages can amplify overall risk.

In practice, the copula-based method in Figures~\ref{fig:side_by_side}(a) and \ref{tab:copula_scenarios} is well-suited for continuous variables and correlation analysis, while the Markov chain method in Figures~\ref{fig:side_by_side}(b) and \ref{tab:markov_chain_scenarios} captures discrete failure modes and transitions more directly. Though these scenarios are hypothetical, they illustrate how each modelling framework can be adapted to real systems. With actual data, parameters and probabilities can be updated, enabling decision-makers to dynamically track how AI adoption affects both the day-to-day performance and potential worst-case outcomes.

\section{Conclusion}

We have demonstrated how a scenario-based framework, incorporating both Markov chains and copulas, can provide a structured way to evaluate AI-induced shifts in multi-stage risk profiles. By integrating lookalike distributions for AI events with classical statistical methods (Monte Carlo simulation, VaR, ES), we see how partial or full AI adoption typically lowers mean outcomes but introduces heavier tails, reflecting the potential for rare but severe failures. The toy results highlight the importance of comparing baseline, partial, and full AI scenarios to understand how improvements in one stage may propagate—and sometimes amplify—downstream risks.

Despite the intuitive nature of these methods, future work must address the computational overhead of simulating large-scale models, as well as the challenge of accurately fitting distributions when AI-specific data are scarce or rapidly evolving. More sophisticated parameter updating (e.g.\ Bayesian or online learning approaches) and research on non-stationary Markov frameworks could better capture emerging AI failure modes. Ultimately, regular updates to model parameters and distributions, fueled by real operational data, will be essential to ensure that scenario modelling remains robust in capturing both day-to-day gains and the tail risks introduced by AI.

\subsection*{Impact Statement}
Our work aims to bring a more quantitative and evidence-based methodology to AI safety, encouraging organizations to evaluate AI’s impact on system-wide risk.

% \newpage
\bibliography{example_paper,refs-risk}
\bibliographystyle{icml2025}

%%%%%%%%%%%%%%%%%%%%%%%%%%%%%%%%%%%%%%%%%%%%%%%%%%%%%%%%%%%%%%%%%%%%%%%%%%%%%%
%%%%%%%%%%%%%%%%%%%%%%%%%%%%%%%%%%%%%%%%%%%%%%%%%%%%%%%%%%%%%%%%%%%%%%%%%%%%%%
% APPENDIX
%%%%%%%%%%%%%%%%%%%%%%%%%%%%%%%%%%%%%%%%%%%%%%%%%%%%%%%%%%%%%%%%%%%%%%%%%%%%%%
%%%%%%%%%%%%%%%%%%%%%%%%%%%%%%%%%%%%%%%%%%%%%%%%%%%%%%%%%%%%%%%%%%%%%%%%%%%%%%
\newpage
\appendix
\onecolumn

\section{Appendix: Lookalike Distribution Simulation}\label{Appendix: Lookalike Distribution Simulation}

A large language model (LLM) can serve as a surrogate data generator or expert oracle for AI-driven sub-events, particularly when real-world AI logs are scarce. Below, we outline a method that uses LLM outputs to validate or refine lookalike distributions for $E_1^{(\mathrm{AI})}$, $E_2^{(\mathrm{AI})}$, and $E_3^{(\mathrm{AI})}$.

\subsection{Setup}
Assign each sub-event (forecasting, picking, delivery) a baseline (non-AI) distribution, and propose an AI-based distribution with heavier tails or shifted means. For example:
\begin{align}
E_1:\,\,&\text{Non-AI} \sim \mathcal{N}(0,\,15^2), 
&&
E_1^{(\mathrm{AI})} \sim (1 - \varepsilon_1)\,\mathcal{N}(0,\,10^2)\;+\;\varepsilon_1 \,\mathrm{Lognormal}(\mu_1,\sigma_1), \\
E_2:\,\,&\text{Non-AI} \sim \Gamma(5,2), 
&&
E_2^{(\mathrm{AI})} \sim (1 - \varepsilon_2)\,\Gamma(8,1)\;+\;\varepsilon_2\,\mathrm{Pareto}(\alpha_2), \\
E_3:\,\,&\text{Non-AI} \sim \mathrm{Lognormal}(1,\,0.5), 
&&
E_3^{(\mathrm{AI})} \sim (1 - \varepsilon_3)\,\mathrm{Lognormal}(0.8,\,0.4)\;+\;\varepsilon_3\,\mathrm{Weibull}(k_3,\lambda_3).
\end{align}
Set initial guesses for $\varepsilon_i$, $\alpha_2$, $k_3$, and $\lambda_3$ via domain expertise or from analogy to mechanical automation data.

\subsection{LLM-Based Sampling.}
Use a prompt to query the LLM for synthetic outcomes that represent the AI’s performance in each sub-event. For forecasting ($E_1$), an example request might be:  
\emph{``Estimate 5 random daily forecast errors (in units) for the AI forecasting model, given it is accurate most of the time but occasionally produces large outliers.''}  
If the LLM returns samples 
\begin{align}
x_1,\,x_2,\,\dots,\,x_m,
\end{align}
treat these as pseudo-observations for $E_1^{(\mathrm{AI})}$. Similarly, obtain samples for $E_2^{(\mathrm{AI})}$ and $E_3^{(\mathrm{AI})}$ using appropriate context or scenario-based prompts.

\subsection{Goodness-of-Fit Check}
Let $F_{\mathrm{AI}}$ denote the current lookalike distribution for a given sub-event. From the LLM’s $m$ samples, compute the empirical CDF 
$
\widehat{F}_{\mathrm{sample}}(x)
$ 
and compare it to 
$
F_{\mathrm{AI}}(x)
$
using, for instance, a Kolmogorov–Smirnov statistic:
\begin{align}
D_{\mathrm{KS}}
=
\sup_{x}
\bigl|\,
\widehat{F}_{\mathrm{sample}}(x)
\;-\;
F_{\mathrm{AI}}(x)
\bigr|.
\end{align}
If $D_{\mathrm{KS}}$ is too large (or a chosen test rejects), adjust parameters to reduce the mismatch.

\subsection{Parameter Updating}
When the AI distribution is parametric (e.g.\ mixture of Gamma and Pareto), re-estimate parameters by maximum likelihood on the LLM sample. For picking times in $E_2$, define
\begin{align}
(\hat{k}, \hat{\theta}, \hat{\alpha_2}, \hat{\varepsilon_2})
=
\mathrm{arg\,max}_{\,k,\theta,\alpha_2,\varepsilon_2}
\prod_{\ell=1}^m 
f_{E_2^{(\mathrm{AI})}}(x_\ell \,\vert\, k,\theta,\alpha_2,\varepsilon_2).
\end{align}
Similarly update $(\varepsilon_1,\mu_1,\sigma_1)$ for $E_1^{(\mathrm{AI})}$ or $(\varepsilon_3,k_3,\lambda_3)$ for $E_3^{(\mathrm{AI})}$. If using a Bayesian approach, treat the LLM sample as observations, update the posterior, and replace point estimates with posterior means or medians.

\subsection{Integrate into Scenario Modelling}
After refining $E_i^{(\mathrm{AI})}$, embed these updated distributions into either a copula or Markov chain dependency structure. For a copula approach, replace $F_{X_i}^{(\mathrm{AI})}$ in the marginals. For a Markov chain approach, adjust transition probabilities or cost/time distributions associated with AI-related states. Then run Monte Carlo simulations to estimate 
$
\Pr(X_{\mathrm{total}}>t),
\mathrm{VaR}_{\alpha},
\mathrm{ES}_{\alpha}
$
for no AI, partial AI, or full AI adoption. If additional LLM samples or real AI data become available, repeat the above steps to iteratively improve the parameter estimates.

\section{Appendix: Mathematical Formalism}
\label{app:detailed_mathematics}

In this appendix, we provide expanded detail on the mathematical formulations used above. We also offer guidance on their implementation, including considerations of computational efficiency and parameter estimation.

\subsection{Markov Chain Monte Carlo Methods}
\label{app:mcmc}

\subsubsection{Discrete-Time Markov Chains and Simulation}
Let $\{X_t\}_{t=0}^{\infty}$ be a discrete-time Markov chain with a finite or countable state space $\mathcal{S}$. We specify a transition matrix $P$, where
\begin{align}
 p_{ij} = \Pr(X_{t+1} = j \,\vert\, X_t = i), \quad \sum_{j \in \mathcal{S}} p_{ij} = 1.
\end{align}
In reliability settings, $\mathcal{S}$ might be states like \emph{Operational, MinorFailure, MajorFailure, Repair}, etc. If we denote the initial distribution by $\alpha$ (a row vector of probabilities), the distribution at time $t$ is $\alpha P^t$.

\emph{Simulation approach.} Given a Markov chain model for an entire process with transitions triggered by sub-event outcomes:
\begin{enumerate}
\item Start in state $X_0$. 
\item For each step $t = 0,1,2,\dots$, sample $X_{t+1}$ from the categorical distribution defined by row $X_t$ of $P$.
\item Track key metrics (time to absorption, number of visits to failure states, etc.).
\end{enumerate}
If we want to incorporate AI, we adjust $P$. For instance, if sub-event $E_i$ is replaced with AI-based technology that changes the probabilities of transitions to failure states, we reflect this by substituting the relevant rows or columns in $P$. The chain might also have more states if AI introduces new categories of failure (e.g., \emph{AI-induced meltdown}).

\subsubsection{Continuous-Time Markov Chains}
Many reliability problems prefer a continuous-time Markov chain (CTMC), where transitions occur at exponential waiting times governed by rates $\lambda_{ij}$. The generator matrix $Q$ has off-diagonal elements $q_{ij} = \lambda_{ij}$ for $j \neq i$ and diagonal elements $q_{ii} = -\sum_{j \neq i} \lambda_{ij}$. Simulation typically involves sampling exponential holding times for each state. This approach is especially relevant for events that occur spontaneously and continuously (e.g., time to breakdown). AI can modify certain rates $\lambda_{ij}$, potentially reducing or increasing the speed of transitions to particular failure states.

\subsubsection{MCMC for Distribution Sampling vs. Markov Risk Models}
One subtlety is the term \emph{Markov Chain Monte Carlo (MCMC)} is often used for sampling from complex high-dimensional distributions, e.g. in Bayesian inference. In the present context, we may do a direct \emph{Monte Carlo} simulation of a Markov chain that represents states of system operation. The distinction is that we do not always need an MCMC algorithm if the chain is of moderate size or can be simulated directly. However, if the chain is extremely large or complicated, advanced MCMC techniques might be employed to approximate its stationary or long-run distribution.

\subsection{Copula Models}
\label{app:copula}

\subsubsection{Sklar’s Theorem}
Copulas are functions that allow us to decompose a joint distribution $F_{X_1,\ldots,X_n}$ into univariate marginals $F_{X_i}$ and a dependence structure $C$. In practice, we often parametrize the copula, such as a Gaussian copula with correlation matrix $\Sigma$ or an Archimedean copula with a parameter $\theta$. For example, for two variables $X_1, X_2$, we have:
\begin{align}
   F_{X_1,X_2}(x_1,x_2) = C\bigl(F_{X_1}(x_1), F_{X_2}(x_2)\bigr).
\end{align}
If $C$ is a Gaussian copula, then
\begin{align}
   C(u_1, u_2; \rho) = \Phi_{\rho}\bigl(\Phi^{-1}(u_1), \Phi^{-1}(u_2)\bigr),
\end{align}
where $\Phi_{\rho}$ is the bivariate normal CDF with correlation $\rho$ and $\Phi^{-1}$ is the inverse of the univariate standard normal. Higher dimensions generalize to correlation matrices $\Sigma$. Other common families include the t-copula, Clayton, Gumbel, and Frank copulas.

\subsubsection{Implementation Steps}
\begin{enumerate}
\item \emph{Marginal fits.} For each sub-event (e.g., time or cost), choose a distribution (Gamma, Weibull, etc.) and estimate parameters from data or domain knowledge. Denote the fitted CDF by $F_{X_i}$.
\item \emph{Transform data to uniforms.} If we have sample data $(x_{i1},\dots,x_{in})$ for sub-events, we convert each $x_{ij}$ to $u_{ij} = F_{X_i}(x_{ij})$. The pairs $(u_{1j},\dots,u_{nj})$ lie in $[0,1]^n$.
\item \emph{Fit copula parameters.} For instance, in a Gaussian copula, we compute the empirical correlation among $\Phi^{-1}(u_{ij})$. For an Archimedean copula, we solve for $\theta$ that best fits the dependence structure.
\item \emph{Simulation.} To generate a new realization:
\begin{enumerate}
   \item Draw $(u_1,\dots,u_n)$ from the copula with estimated parameters,
   \item Invert each via $x_i = F_{X_i}^{-1}(u_i)$.
\end{enumerate}
\end{enumerate}
When AI modifies sub-event $i$, we simply replace the marginal CDF $F_{X_i}$ with the AI version and possibly re-estimate the copula parameters if the dependence structure is believed to change.

\end{document}